\documentclass[10pt,twocolumn,letterpaper]{article}

\usepackage{cvpr}
\usepackage{times}
\usepackage{epsfig}
\usepackage{graphicx}
\usepackage{amsmath}
\usepackage{amssymb}
\usepackage{algorithm}
\usepackage{algorithmic}
\usepackage{multirow}
\usepackage{diagbox}

\usepackage[pagebackref=true,breaklinks=true,letterpaper=true,colorlinks,bookmarks=false]{hyperref}

\cvprfinalcopy 

\def\cvprPaperID{0602} 

\ifcvprfinal
\fi
\begin{document}
	
	\title{TubeTK: Adopting Tubes to Track Multi-Object in a One-Step Training Model}
	
	\author{Bo Pang, Yizhuo Li, Yifan Zhang, Muchen Li$^\dag$, Cewu Lu\footnotemark[1]\\
		Shanghai Jiao Tong University, \dag Huazhong University of Science and Technology\\
		{\tt\small \{pangbo, liyizhuo, zhangyf\_sjtu\, lucewu\}@sjtu.edu.cn, muchenli@alumni.hust.edu.cn}}
	
	\maketitle
	\renewcommand{\thefootnote}{\fnsymbol{footnote}}
	\footnotetext[1]{Cewu Lu is the corresponding author, member of Qing Yuan Research Institute and MoE Key Lab of Artificial Intelligence, AI Institute, Shanghai Jiao Tong University, China.}
	
	\thispagestyle{empty}
	\begin{abstract}
		\vspace{-0.1in}
		Multi-object tracking is a fundamental vision problem that has been studied for a long time. As deep learning brings excellent performances to object detection algorithms, Tracking by Detection (TBD) has become the mainstream tracking framework. Despite the success of TBD, this two-step method is too complicated to train in an end-to-end manner and induces many challenges as well, such as insufficient exploration of video spatial-temporal information, vulnerability when facing object occlusion, and excessive reliance on detection results. To address these challenges, we propose a concise end-to-end model \textbf{TubeTK} which only needs one step training by introducing the ``bounding-tube" to indicate temporal-spatial locations of objects in a short video clip. TubeTK provides a novel direction of multi-object tracking, and we demonstrate its potential to solve the above challenges without bells and whistles.
		We analyze the performance of TubeTK on several MOT benchmarks and provide empirical evidence to show that TubeTK has the ability to overcome occlusions to some extent without any ancillary technologies like Re-ID. Compared with other methods that adopt private detection results, our one-stage end-to-end model achieves state-of-the-art performances even if it adopts no ready-made detection results. We hope that the proposed TubeTK model can serve as a simple but strong alternative for video-based MOT task. The code and models are available at \url{https://github.com/BoPang1996/TubeTK}. 
	\end{abstract}
	
	\begin{figure}
		\begin{center}
			\includegraphics[width=\linewidth]{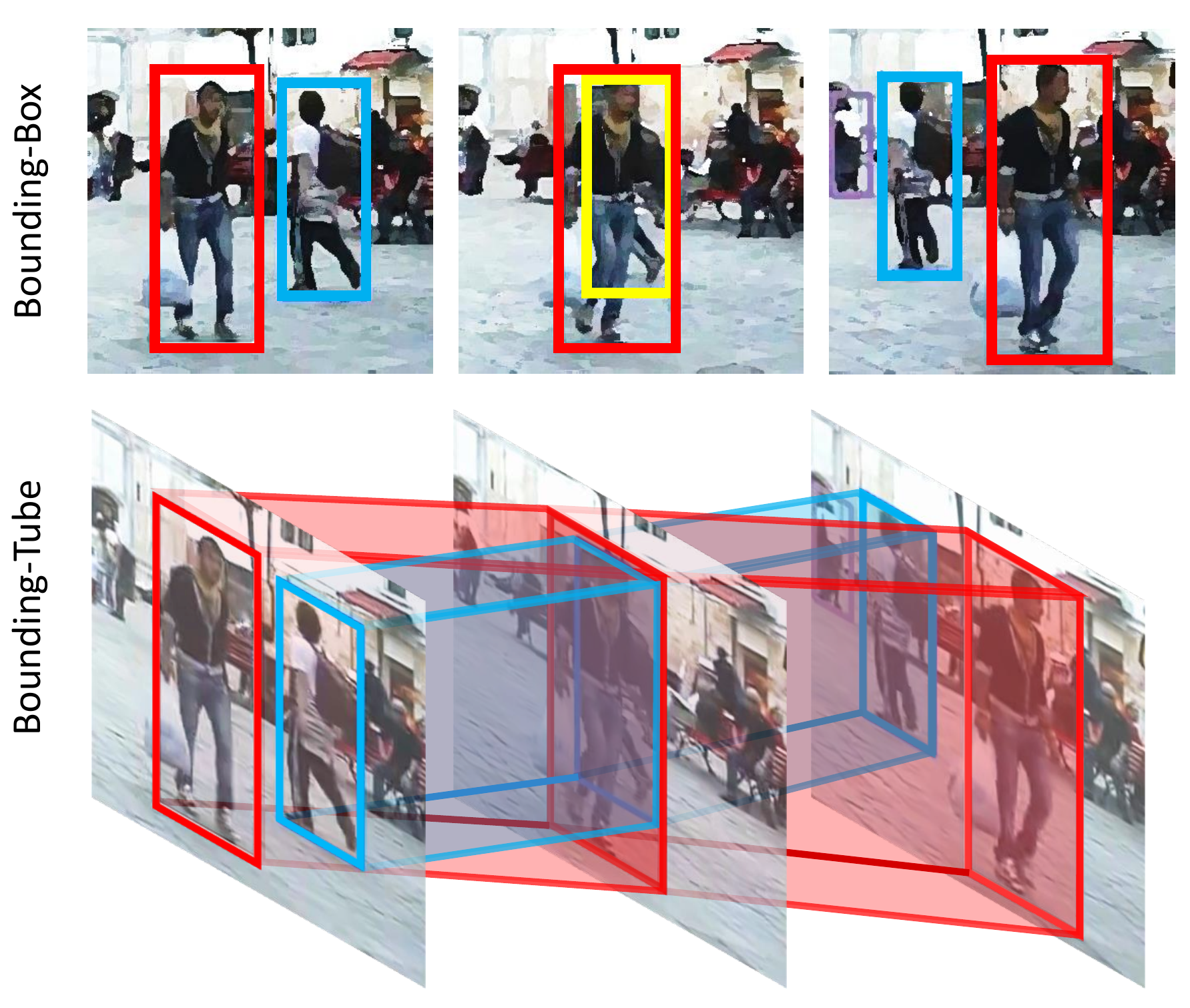}
		\end{center}
		\vspace{-0.1in}
		\caption{Bounding-boxes and bounding-tubes. As shown in the first row, it is difficult to detect the severely occluded target (the yellow box)  by the spatial box without temporal information. In our TubeTK (the second row), it generates bounding-tubes based on temporal-spatial features that encode targets' spatial location and moving trail at the same time. This leads to a one-step training tracking method which is more robust when facing occlusions.}
		\label{fig:tube_sketch}
		\vspace{-0.1in}
	\end{figure}
	
	\vspace{-0.23in}
	\section{Introduction}
	\vspace{-0.07in}
	Video multi-object tracking (MOT) is a fundamental yet challenging task that has been studied for a long time. It requires the algorithm to predict the temporal and spatial location of objects and classify them into correct categories. The current mainstream trackers such as~\cite{yu2016poi,bergmann2019tracking,chu2019famnet,bae2017confidence,fagot2016improving} all adopt the tracking-by-detection (TBD) framework. As a two-step method, this framework simplifies the tracking problem into two parts: detecting the spatial location of objects and matching them in the temporal dimension. Although this is a successful framework, it is important to note that TBD method suffers from some drawbacks:
	\begin{enumerate}
		\vspace{-0.1in}
		\item As shown in \cite{yu2016poi,fang2018recurrent}, the performances of models adopting TBD framework dramatically vary with detection models. This excessive reliance on image detection results limits performances of the MOT task. Although there are some existing works aiming at integrating the two steps more closely~\cite{zhang2018integrated,feichtenhofer2017detect,bergmann2019tracking}, the problems are still not solved fundamentally because of the relatively independent detection model.
		
		\vspace{-0.1in}
		\item  Due to image-based detection models employed by TBD, the tracking models are weak when facing severe object occlusions (see Fig.~\ref{fig:tube_sketch}). It is extremely difficult to detect occluded objects only through spatial representations~\cite{bergmann2019tracking}. The low quality detection further makes tracking unstable, which leads to more complicated design of matching mechanism~\cite{sun2019deep,voigtlaender2019mots}.  
		
		\vspace{-0.1in}
		\item  As a video level task, MOT requires models to process spatial-temporal information (STI) integrally and effectively. To some extent, the above problems are caused by the separate exploration of STI: detectors mainly model spatial features and trackers capture temporal ones~\cite{sadeghian2017tracking,chu2019famnet,fang2018recurrent,sun2019deep}, which casts away the semantic consistency of video features and results in incomplete STI at each step.		
		\vspace{-0.05in}
	\end{enumerate}
	
	Nowadays, many video tasks can be solved in a simple one-step end-to-end method such as the I3D model~\cite{carreira2017quo} for action recognition~\cite{li2019hake}, TRN~\cite{zhou2018temporal} for video relational reasoning, and MCNet~\cite{villegas2017decomposing} for video future prediction. As one of the fundamental vision tasks, MOT still does not work in a simple elegant method and the drawbacks of TBD mentioned above require assistance of some other techniques like Re-ID~\cite{bergmann2019tracking,long2018real}. It is natural to ask a question: \textit{Can we solve the multi-object tracking in a neat one-step framework?} In this way, MOT can be solved as a stand-alone task, without restrictions from detection models. We answer it in the affirmative and for the first time, we demonstrate that the much simpler one-step tracker even achieves better performance than the TBD-based counterparts.
	
	In this paper, we propose the \textbf{TubeTK} which conducts the MOT task by regressing the bounding-tubes (Btubes) in a 3D manner. Different from 3D point-cloud~\cite{you2018prin}, this 3D means 2D spatial and 1D temporal dimensions. As shown in Fig.~\ref{fig:tube_sketch}, a Btube is defined by 15 points in space-time compared to the traditional 2D box of 4 points. Besides the spatial location of targets, it also captures the temporal position. More importantly, the Btube encodes targets' motion trail as well, which is exactly what MOT needs. Thus, Btubes can well handle spatial-temporal information integrally and largely bridge the gap between detection and tracking.
	
	To predict the Btube that captures spatial-temporal information, we employ a 3D CNN framework. By treating a video as 3D data instead of a group of 2D image frames, it can extract spatial-temporal features simultaneously. This is a more powerful and fully automatic method to extract tracking features, where the handcrafted features such as optical flow~\cite{simonyan2014two}, segmentation~\cite{voigtlaender2019mots,fang2019instaboost,xu2018srda}, human pose~\cite{fang2018learning,fang2017rmpe,Wang_2019_ICCV} or targets interactions~\cite{sadeghian2017tracking,li2018transferable,fang2018pairwise,qi2018learning} are not needed. The network structure is inspired by recent advances of one-stage anchor-free detectors~\cite{tian2019fcos,duan2019centernet} where the FPN~\cite{lin2017feature} is adopted to better track targets of different scales and the regression head directly generates Btubes. After that, simple IoU-based post-processing is applied to link Btubes and form final tracks. The whole pipeline is made up of fully convolutional networks and we show the potential of this compact model to be a new tracking paradigm. 
	
	\noindent
	The proposed TubeTK enjoys the following advantages:
	\begin{enumerate}
		\vspace{-0.1in}
		\item With TubeTK, MOT now can be solved by a simple one-step training method as other video tasks. Without constraint from detection models, assisting technologies, and handcrafted features,
		TubeTK is considerably simpler when being applied and it also enjoys great potential in future research.
		
		\vspace{-0.1in}
		\item TubeTK adequately extracts spatial-temporal features simultaneously and these features capture information of motion tendencies. Thus, TubeTK is more robust when faced with occlusions.
		\vspace{-0.1in}
		\item Without bells and whistles, the end-to-end-trained TubeTK achieves better performances compared with TBD-based methods on MOT15, 16, and 17 dataset~\cite{leal2015motchallenge,milan2016mot16}. And we show that the Btube-based tracks are smoother (fewer FN and IDS) than the ones based on pre-generated image-level bounding-boxes. 
	\end{enumerate}

	\section{Related Work}
	\vspace{-0.05in}
	\paragraph{Tracking-by-detection-based model}
	Research based on the TBD framework often adopts detection results given by external object detectors~\cite{redmon2016you,liu2016ssd,lu2018beyond} and focuses on the tracking part to associate the detection boxes across frames. Many associating methods have been utilized on tracking models. In~\cite{berclaz2011multiple,jiang2007linear,zhang2008global,pirsiavash2011globally,lenz2015followme}, every detected bounding-box is treated as a node of graph, the associating task is equivalent to determining the edges where maximum flow~\cite{berclaz2011multiple,xiu2018poseflow}, or equivalently, minimum cost~\cite{pirsiavash2011globally,jiang2007linear,zhang2008global} are usually adopted as the principles. Recently, with the development of deep learning, appearance-based matching algorithms have been proposed~\cite{kim2015multiple,sadeghian2017tracking,fang2018recurrent}. By matching targets with similar appearances such as clothes and body types, models can associate them over long temporal distances. Re-ID techniques~\cite{kuo2011does,bergmann2019tracking,tang2017multiple} are usually employed as an auxiliary in this matching framework. 
	\begin{figure*}
		\begin{center}
			\includegraphics[width=\linewidth, height=2.5in]{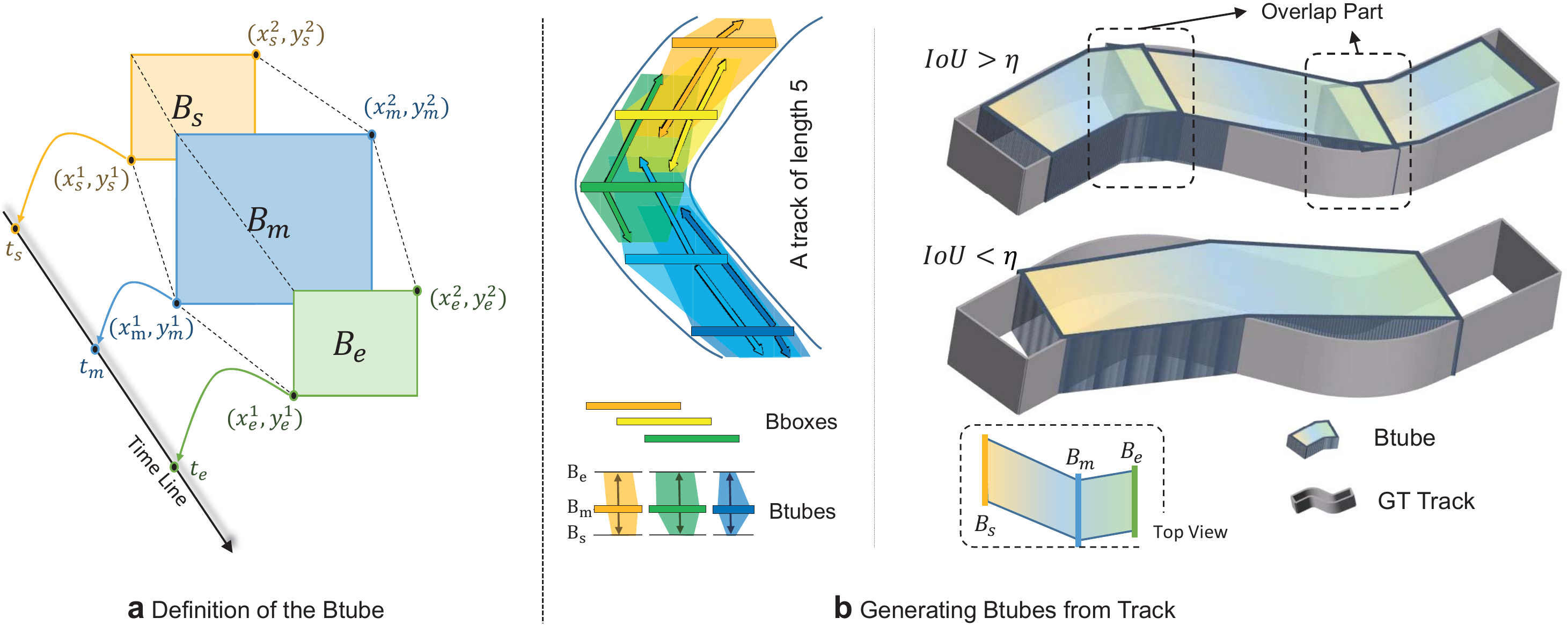}
		\end{center}
		\vspace{-0.15in}
		\caption{Definition and generation of the Btube. \textbf{a}: A Btube can be seen as the combination of three bounding-boxes $B_s$, $B_m$, and $B_e$ from different video frames. A Btube has 15 degrees of freedom, which can be determined by the spatial locations of the three bounding-boxes (4$\times$3 degrees) and their temporal positions (3 degrees, $t_s$, $t_m$, and $t_e$). \textbf{b}: Btubes are generated from whole tracks. Left: For each bounding-box in a track, we treat it as the $B_m$ of one Btube then look forward and backward to find its $B_e$ and $B_s$ in the track. Right: A longer Btube can capture more temporal features but the IoU between it and the track is lower ($\eta$ is the IoU threshold), which leads to bad moving trails as the second row shows. Overlaps between the Btubes are used for linking them.}
		\label{fig:tube_def}
		\vspace{-0.15in}
	\end{figure*}
	
	\vspace{-0.17in}
	\paragraph{Bridging the gap between detection and tracking}
	Performances of image-based object detectors are limited when facing dense crowds and serious occlusions. Thus, some works try to utilize extra information such as motion~\cite{sadeghian2017tracking} or temporal features learned by the track step to aid detection. One simple direction is to add bounding-boxes generated by the tracking step into the detection step~\cite{long2018real,chu2017online}, but this does not affect the original detection process. In \cite{zhang2018integrated}, the tracking step can efficiently improve the performance of detection by controlling the NMS process. \cite{feichtenhofer2017detect} proposes a unified CNN structure to jointly perform detection and tracking tasks. By sharing features and conducting multi-task learning, it can further reduce the isolation between the two steps. The authors of~\cite{wang2019torwards} propose a joint detection and embedding framework where the detection and associating steps share same features. 
	Despite these works' effort to bridge the gap between detection and tracking,
	they still treat them as two separate tasks and can not well utilize spatial-temporal information.
	
	\vspace{-0.17in}
	\paragraph{Tracking framework based on trajectories or tubes}
	Tubes can successfully capture motion trails of targets, which are important for tracking. There are previous works that adopt tubes to conduct MOT or video detection~\cite{shao2018find} tasks. In~\cite{kang2016object,kang2017t}, a tubelet proposal module combining detection results into tubes is adopted to solve the video detection task. And \cite{zhu2018online} employs a single-object tracking method to capture subjects' trajectories. Although these works propose and utilize the concept of tubes, they still utilize external detection results and form tubes at the second step, instead of directly regressing them. Thus they are still TBD methods and the problems stated above are not solved.
	
	\begin{figure*}
		\begin{center}
			\includegraphics[width=\linewidth, height=3.0in]{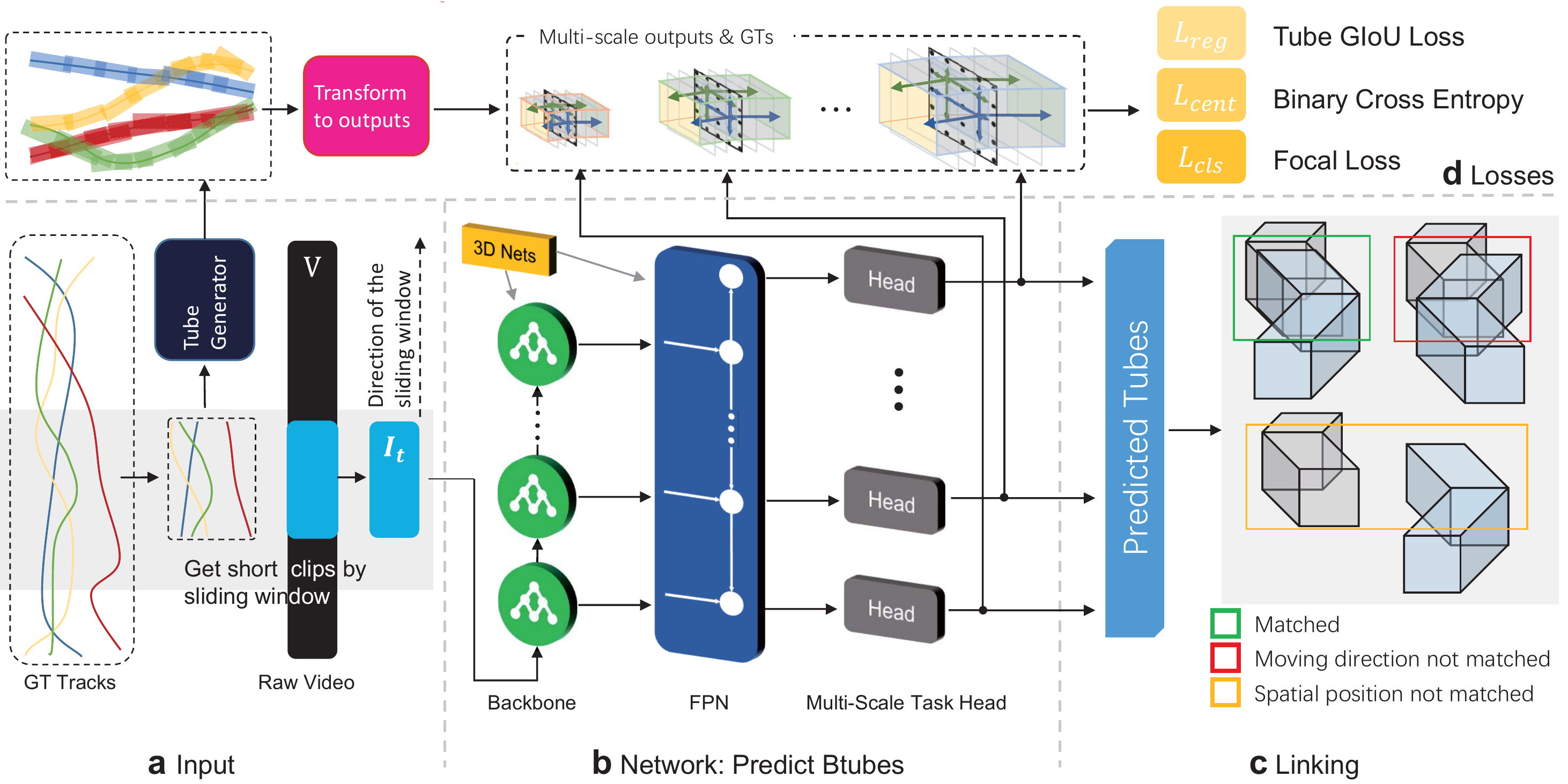}
		\end{center}
		\vspace{-0.1in}
		\caption{The pipeline of our TubeTK. \textbf{a}: Given a video $V$ and the corresponding ground-truth tracks, we cut them into short clips in a sliding window manner to get inputs of the network. \textbf{b}: To model spatial-temporal information in video clips, we adopt 3D convolutional layers to build our network which consists of a backbone, an FPN, and a few multi-scale heads. Following FCOS~\cite{tian2019fcos}, the multi-scale heads are responsible for targets with different scales respectively. The 3D network directly predicts Btubes. \textbf{c}: We link the predicted Btubes that have the same spatial positions and moving directions in the overlap part into whole tracks. \textbf{d}: In the training phase, the GT tracks are split into Btubes and then they are transformed into the same form of the network's output: target maps (see Fig.~\ref{fig:output} for details). The target and predicted maps are fed into three loss functions to train the model: the Focal loss for classifying the foreground and background, BCE for giving out the center-ness, and GIoU loss for regressing Btubes.}
		\label{fig:pipeline}
		\vspace{-0.1in}
	\end{figure*}
	
	\vspace{-0.05in}
	\section{The Proposed Tracking Model}
	\vspace{-0.05in}
	We propose a new one-step end-to-end training MOT paradigm, the TubeTK. Compared with the TBD framework, this paradigm can better model spatial-temporal features and alleviate problems led by dense crowds and occlusions. In this section, we will introduce the entire pipeline in the following arrangement: 1) We first define the Btube which is a 3D extension of Bbox and introduce its generation method in Sec.~\ref{sec:btube_def}. 2) In Sec.~\ref{sec:network}, we introduce the deep network adopted to predict Btubes from input videos. 3) Next, we interpret the training method tailored for Btubes in Sec.~\ref{sec:training}. 4) Finally, we propose the parameter-free post-processing method to link the predicted Btubes in Sec.~\ref{sec:linking}.
	
	\vspace{-0.05in}
	\subsection{From Bounding-Box to Bounding-Tube}\label{sec:btube_def}
	\vspace{-0.05in}
	Traditional image-based bounding-box (Bbox) which serves as the smallest enclosing box of a target can only indicate its spatial position, while for MOT, the pattern of targets' temporal positions and moving directions is of equal importance. Thus, we go down to consider how can we extend the bounding-box to simultaneously represent the temporal position and motion, with which, models can overcome occlusions shorter than the receptive field.
	
	\vspace{-0.2in}
	\paragraph{Btube definition}
	Adopting a 3D Bbox to point out an object across frames is the simplest extension method, but obviously, this 3D Bbox is too sparse to precisely represent the target's moving trajectory. Inspired by the tubelet in video detection task~\cite{kang2016object,kang2017t}, we design a simplified version, called bounding-tube (Btube), for the dimension of original tubelets is too large to directly regress. A Btube can be uniquely identified in space and time by 15 coordinate values and it is generated by a method similar to the linear spline interpolation which splits a whole track into several overlapping Btubes. 
	
	As shown in Fig.~\ref{fig:tube_def} \textbf{a}, a Btube $T$ is a decahedron composed of 3 Bboxes in different video frames, namely $B_s$, $B_m$, and $B_e$, which need 12 coordinate values to define. And 3 other values are used to point out their temporal positions. This setting allows the target to change its moving direction once in a short time. Moreover, its length-width ratio can change linearly, which makes the Btube more robust when facing pose and scale changes led by perspective. By interpolation between $(B_s, B_m)$ and $(B_m, B_e)$, we can restore all the bounding-boxes $\{B_s, B_{s+1},..., B_m, ..., B_{e-1}, B_{e}\}$ that constitute the Btube. Note that $B_m$ does not have to be exactly at the midpoint of $B_s$ and $B_e$. It may be closer to one of them. Btubes are designed to encode spatial and temporal information simultaneously. It can even reflect targets' moving trends which are important in MOT task. These specialties make Btubes contain much more useful semantics than traditional Bboxes.
	
	\vspace{-0.2in}
	\paragraph{Generating Btubes from tracks}
	Btubes can only capture simple linear trajectories, thus we need to disassemble complex target's tracks into short clips, in which motions can approximately be seen as linear and captured by our Btubes.
	
	The disassembly process is shown in Fig.~\ref{fig:tube_def} \textbf{b}. We split a whole track into multiple overlapping Btubes by extending \textbf{EVERY} Bbox in it to a Btube. We treat each Bbox as the $B_m$ of one Btube then look forward and backward in the track to find its corresponding $B_e$ and $B_s$. We can extend Bboxes to longer Btubes for capturing more temporal information, but long Btubes generated by linear interpolation cannot well represent the complex moving trail (see Fig.~\ref{fig:tube_def}). To balance this trade-off, we set each Btube to be the longest one which satisfies that the mean IoU between its interpolated bounding-boxes $B$ and the ground-truth bounding-boxes $B^\star$ is no less than the threshold $\eta$:
	
	\vspace{-0.11in}
	\begin{equation}
	\small
	\begin{aligned}
	\max ~~~~~& e - s   \\
	\text{s.t.} ~~~~~& {\rm mean}(\{{\rm IoU}(B_i, B_i^\star)\}) \geq \eta\\
	& i \in \{s, s+1, ..., m,...,e\}
	\end{aligned}
	\end{equation}
	\vspace{-0.11in}
	
	This principle allows to dynamically generate Btubes with different lengths. When the moving trajectory is monotonous, the Btubes will be longer to capture more temporal information. While when the motion varies sharply, it will generate shorter Btubes to better fit the trail.
	
	\vspace{-0.2in}
	\paragraph{Overcoming the occlusion}
	Btubes guide models to capture moving trends. Thus, when facing occlusions, these trends will assist in predicting the position of shortly invisible targets. Moreover, this specialty can reduce the ID switches at the crossover point of two tracks because two crossing tracks trend to have different moving directions.
	
	\vspace{-0.05in}
	\subsection{Model Structure}
	\vspace{-0.05in}
	\label{sec:network}
	With Btubes that encode the spatial-temporal position, we can handle the MOT task in one step learning without the help of external object detectors or handcrafted matching features. To fit Btubes, we adopt the 3D convolutional structure~\cite{ji20123d} to capture spatial-temporal features, which is widely used in the video action recognition task~\cite{carreira2017quo,hara2017learning,feichtenhofer2018slowfast}. The whole pipeline is shown in Fig.~\ref{fig:pipeline}.
	
	\vspace{-0.2in}
	\paragraph{Network structure}
	The network consists of a backbone, an FPN~\cite{lin2017feature}, and a few multi-scale task heads.
	
	Given a video $V \in \mathbb{R}^{T,H,W,C}$ to track, where $T$, $H$, $W$  and $C=3$ are frame number, height, width, and input channel respectively, we split it into short clips $I_t$ as inputs. $I_t$ starts from frame $t$ and its length is $l$. As Btubes are usually short, the split clips can provide enough temporal information and reduce the computational complexity. Moreover, by adopting a sliding window scheme, the model can work in an online manner. 
	The 3D-ResNet~\cite{hara2018can,he2016deep} is applied as the backbone to extract the basic spatial-temporal feature groups $\{G^i\}$ with multiple scales. $i$ denotes the level of the features which are generated by stage $i$ of 3D-ResNet. Like the RetinaNet~\cite{lin2017focal} and FCOS~\cite{tian2019fcos}, a 3D version FPN in which the 2D-CNN layers are simply replaced by 3D-CNNs~\cite{ji20123d} then takes $\{G^i\}$ as input and outputs multi-scale feature map groups $\{F^i\}$. This multi-scale setting can better capture targets with different scales. For each $F^i$, there is a task head composed of several CNN layers to output regressed Btubes and confidence scores. This fully 3D network processes temporal-spatial information simultaneously, making it possible to extract more efficient features.
	
	\begin{figure}
		\begin{center}
			\includegraphics[width=\linewidth]{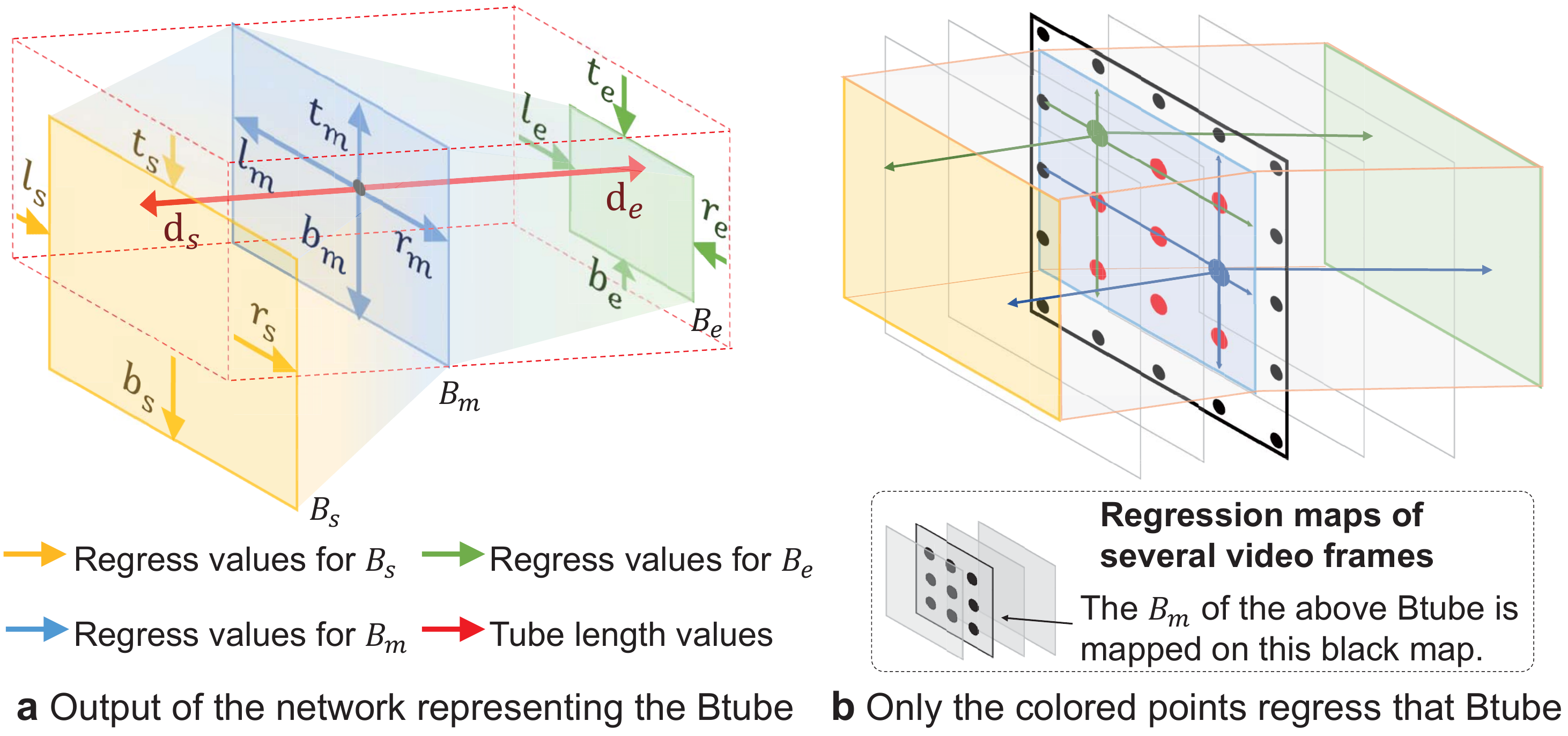}
		\end{center}
		\vspace{-0.15in}
		\caption{Regression method and the matchup between output maps and GT Btubes. \textbf{a}: The model is required to regress the relative temporal and spatial position to focus on moving patterns. \textbf{b}: Each Btube can be regressed by several points in the output map. The colored points on the black map are inside the Btube's $B_m$, so they are responsible for this Btube. Even through on the grey maps, there are some points also inside the Btube, they do not predict it because they are not on its $B_m$.}
		\label{fig:output}
		\vspace{-0.2in}
	\end{figure}
	
	\vspace{-0.2in}
	\paragraph{Outputs}
	Each task head generates three output maps: the confidence map, regression map, and center-ness map following FCOS~\cite{tian2019fcos}. The center-ness map is utilized as a weight mask applied on the confidence map in order to reduce confidence scores of off-center boxes. The sizes of these three maps are the same. Each point $p$ in the map can be mapped back to the original input image. If the corresponding point of $p$ in the original input image is inside the $B_m$ of one Btube, then $p$ will regress its position (see Fig.~\ref{fig:output}). With $p$ the Btube position $\mathbf{r}$ can be regressed by 14 values: four for $B_m$ $\{l_m, t_m, r_m, b_m\}$,  four for $B_s$ $\{l_s, t_s, r_s, b_s\}$, four for $B_e$ $\{l_e, t_e, r_e, b_e\}$, and two for the tube length $\{d_s, d_e\}$. Their definitions are shown in Fig.~\ref{fig:output}. We utilize relative distances with respect to $B_m$, instead of absolute ones, to regress Btubes aiming to make the model focus on moving trails. The center-ness $\mathbf{c}$ which servers as the weighting coefficient of confidence score $\mathbf{s}$ is defined as:
	
	\vspace{-0.15in}
	\begin{equation}\label{eq:centerness}
	\small
	\mathbf{c} = \sqrt{\frac{\min{l_m, r_m}}{\max{l_m, r_m}}\times \frac{\min{t_m, b_m}}{\max{t_m, b_m}}\times\frac{\min{d_s, d_e}}{\max{d_s, d_e}}}
	\vspace{-0.05in}
	\end{equation}
	Although $\mathbf{c}$ can be calculated directly from the predicted $\mathbf{r}$, we adopt a head to regress it, and $\mathbf{c}^\star$ calculated based on GT $\mathbf{r}^\star$ by Eq.~\ref{eq:centerness} is utilized as the ground-truth to train the head. 
	
	Following the FCOS~\cite{tian2019fcos}, different task heads are responsible for detecting objects within a range of different sizes respectively, which can largely alleviate the ambiguity caused by one point $p$ falling into multiple Btubes' $B_m$.
	
	\vspace{-0.05in}
	\subsection{Training Method}~\label{sec:training}
\begin{figure}
		\begin{center}
			\includegraphics[width=\linewidth]{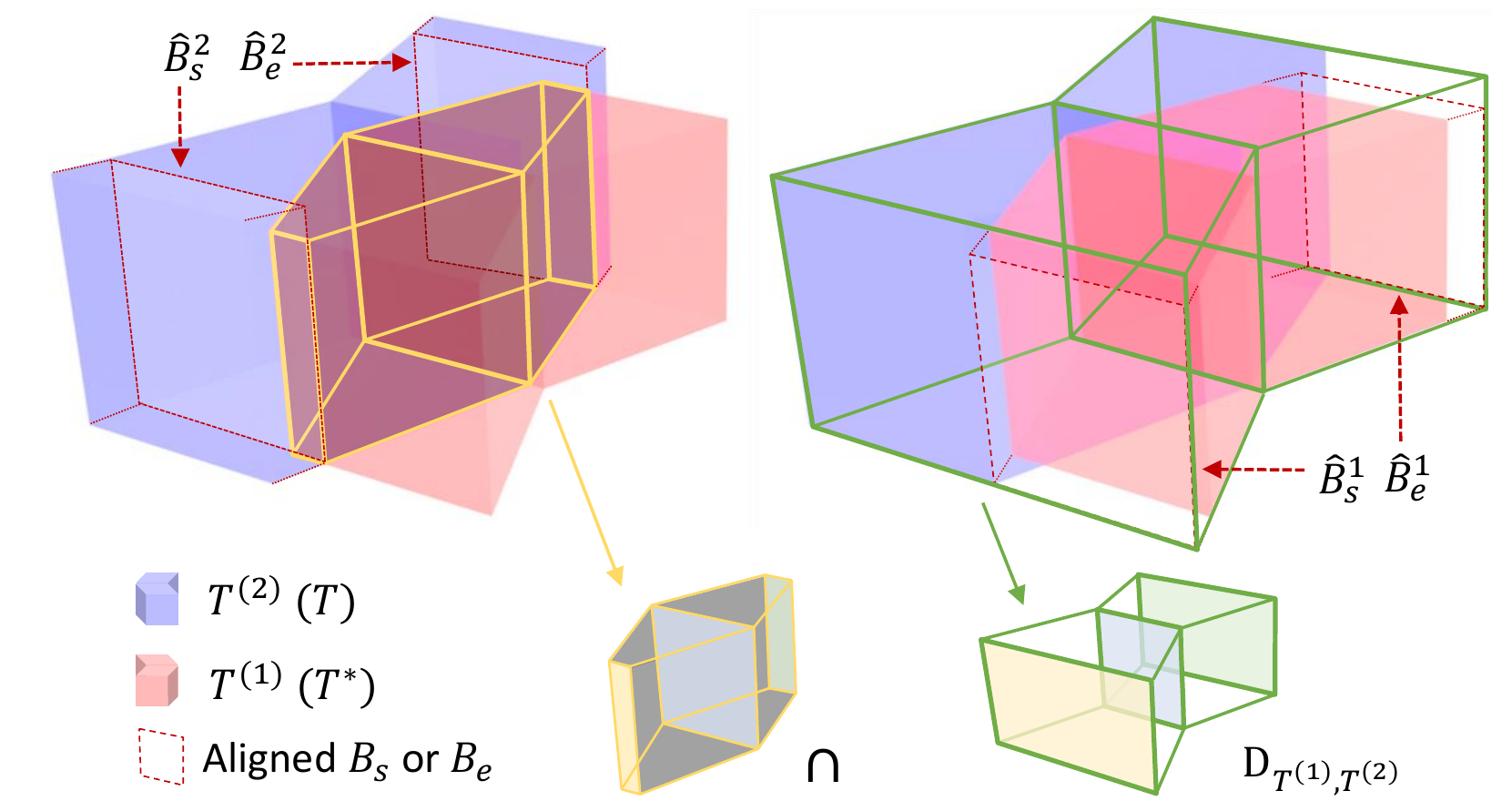}
		\end{center}
		\vspace{-0.15in}
		\caption{Visualization of the calculation process of Tube GIoU. The intersection and $D_{T,T^*}$ of targets are also decahedrons, thus the volume of them can be calculated in the same way of Btubes.}
		\label{fig:giou}
		\vspace{-0.15in}
	\end{figure}

	\vspace{-0.45in}
	\paragraph{Tube GIoU}
	IoU is the most popular indicator to evaluate the quality of the predicted Bbox, and it is usually used as the loss function. GIoU~\cite{rezatofighi2019generalized} loss is an extension of IoU loss which solves the problem that there is no supervisory information when the predicted Bbox has no intersection with the ground truth. GIoU of Bbox is defined as:
	\vspace{-0.1in}
	\begin{equation}
	\small
	{\rm GIoU}(B, B^\star) = {\rm IoU}(B, B^\star) - \frac{|D_{B, B^\star}\backslash (B\cup B^\star)|}{|D_{B, B^\star}|}
	\vspace{-0.1in}
	\end{equation}
	
	where $D_{B, B^\star}$ is the smallest enclosing convex object of $B$ and $B^\star$. We extend the definition of GIoU to make it compatible with Btubes. According to our regression method, $B_m$ and $B_m^\star$ must be on the same video frame, which makes the calculation of BTube's volume, intersection $\bigcap$ and smallest tube enclosing object $D_{T, T^\star}$ straightforward. As shown in Fig.~\ref{fig:giou}, we can treat each Btube as two square frustums sharing the same underside. Because $B_m$ and $B_m^\star$ are on the same video frame, $\bigcap$ and $D_{T, T^\star}$ are also composed of two adjoining square frustums whose volumes are easy to calculate (Detail algorithm is shown in supplementary files). Tube GIoU and Tube IoU are the volume extended version of the original area ones. 
	
	\vspace{-0.2in}
	\paragraph{Loss function}
	For each point $p$ in map $M$, we denote its confidence score, regression result, and center-ness as $\mathbf{s}_p$, $\mathbf{r}_p$, and $\mathbf{c}_p$. The training loss function can be formulated as:
	
	\vspace{-0.25in}
	\begin{equation}
	\vspace{-0.1in}
	\small
	\begin{aligned}
	L(\{\mathbf{s}_p\}, \{\mathbf{r}_p\}, \{\mathbf{c}_p\})= & \frac{1}{N_{pos}}\sum_{p\in M}{L_{cls}(\mathbf{s}_p, \mathbf{s}^\star_p)}\\
	+ & \frac{\lambda}{N_{pos}}\sum_{p\in M}{L_{reg}(\mathbf{r}_p, \mathbf{r}^\star_p)}\\
	+ & \frac{\alpha}{N_{pos}}\sum_{p\in M}{L_{cent}(\mathbf{c}_p, \mathbf{c}^\star_p)}\\
	\end{aligned}
	\end{equation}
	where $\star$ denotes the corresponding ground truth. $N_{pos}$ denotes the number of positive foreground samples. $\lambda$ and $\alpha$ are the weight coefficients which are assigned as 1 in the experiments. $L_{cls}$ is the focal loss proposed in \cite{lin2017focal}, $L_{cent}$ is the binary cross-entropy loss, and $L_{reg}$ is the Tube GIoU loss which can be formulated as:
	\vspace{-0.06in}
	\begin{equation}
	\vspace{-0.06in}
	\small
	L_{reg}(\mathbf{r}_p, \mathbf{r}^\star_p) =  \mathbb{I}_{\{\mathbf{s}^\star_p = 1\}}(1-{\rm TGIoU}(\mathbf{r}_p, \mathbf{r}^\star_p))
	\end{equation}
	where $\mathbb{I}_{\{\mathbf{s}^\star_p = 1\}}$ is the indicator function, being 1 if $\mathbf{s}^\star_p = 1$ and 0 otherwise. $\rm TGIoU$ is the Tube GIoU.
	
	\vspace{-0.05in}
	\subsection{Linking the Bounding-Tubes}~\label{sec:linking}
	After getting predicted Btubes, we only need an IoU-based method without any trainable parameters to link them into whole tracks.
	\vspace{-0.2in}
	\paragraph{Tube NMS}
	Before the linking principles, we will first introduce the NMS method tailored for Btubes. As Btubes are in 3D space, if we conduct a pure 3D NMS, the huge number of them will lead to large computational overhead. Thus, we simplify the 3D NMS into a modified 2D version. The NMS operation is only conducted among the Btubes whose $B_m$ is on the same video frame. Traditional NMS eliminates targets that have large IoU. However, this method will break at least one track when two or more tracks intersect. Due to the temporal information encoded in Btubes, we can utilize $B_s$ and $B_e$ to perceive the moving direction of targets. Often the directions of intersecting tracks are different, thus the IoU of their $B_s$, $B_m$, and $B_e$ will not all be large. In the original NMS algorithm, it will suppress one of two Btubes with IoU larger than the threshold $\gamma$, while in the Tube NMS, we set two thresholds $\gamma_1$ and $\gamma_2$, and for two Btubes $T^{(1)}$ and $T^{(2)}$, suppression is conduct when 
	${\rm IoU}(B_m^{(1)}, B_m^{(2)}) > \gamma_1 ~\&~ {\rm IoU}(B_{s'}^{(1)}, B_{s'}^{(2)}) > \gamma_2 ~\&~ {\rm IoU}(B_{e'}^{(1)}, B_{e'}^{(2)}) > \gamma_2$, where $s' = \max(s^{(1)}, s^{(2)})$, $e' = \min(e^{(1)}, e^{(2)})$ and $B_{s'}$ is generated by interpolation. 
	
	\vspace{-0.2in}
	\paragraph{Linking principles}
	After the Tube NMS pre-processing, we need to link all the rest Btubes into whole tracks. The linking method is pretty simple which is only an IoU-based greedy algorithm without any learnable parameters or assisting techniques like appearance matching or Re-ID. 
	
	Due to the overlap of Btubes in the temporal dimension, we can focus on it to calculate the frame-based IoU for linking. Given a track $K_{(s_1, e_1)}$ starting from frame $s_1$ and ending at frame $e_1$, and a Btube $T_{(s_2, e_2)}$, we first find the overlap part: $O_{(s_3, e_3)}$ where $s_3 = \max(s_1, s_2)$ and $e_3 = \min(e_1, e_2)$. If $s_3 > e_3$, $K$ and $T$ have no overlap and do not need to link. When they are overlapping, we calculate the matching score $\mathcal{M}$ as: 
	\vspace{-0.06in}
	\begin{equation}
	\small
	\mathcal{M}(K, T) = [\sum_{f\in O}{{\rm IoU}(K_f, T_f)}]/|O|
	\vspace{-0.06in}
	\end{equation}
	where $K_f$ and $T_f$ denote the (interpolated) bounding-boxes at frame $f$ in $K$ and $T$. $|O|$ is the number of frames in $O$. If $\mathcal{M}$ is larger than the linking threshold $\beta$, we link them by adding the interpolated bounding-boxes of $T$ onto $K$. It should be noted that in the overlap part, we average the bounding-boxes from $T$ and $K$ to reduce the deviation caused by the linear interpolation. The linking function can be formulated as:
	\vspace{-0.06in}
	\begin{equation} \label{eq:link}
	\vspace{-0.06in}
	\small
	\begin{aligned}
	K^{new} &= {\rm Link}(K_{(s_1, e_1)}, T_{(s_2, e_2)}) \\
	&= K_{(s_1, s_3)} + {\rm Avg}(K_{(s_3, e_3)}, T_{(s_3, e_3)}) + T_{(e_3, e_2)}
	\end{aligned}
	\end{equation}
	where we assume that $e_1 < e_2$, and $+$ denotes jointing two Btubes (or tracks) without overlap.
	
	To avoid ID switch at intersection of two tracks, we also take moving directions into account. The moving direction vector (MDV) of a Btube (or track) starts from the center of its $B_s$ and ends at $B_e$'s center. We hope the track and Btube with similar directions can be more likely to link. 
	Thus, we compute the angle $\theta$ between the MDV of $T_{(s_3, e_3)}$ and $K_{(s_3, e_3)}$ and take $\cos\theta$ as a weighted coefficient masked on $\mathcal{M}$ to adjust the matching score.
	The final matching score utilized to link is $\mathcal{M}' = \mathcal{M} * (1 + \phi*\cos{\theta})$, where $\phi > 0$ is a hyper-parameter. If the direction vectors of the track and Btube form an acute angle, $\cos{\theta} > 0$ and their matching score $\mathcal{M}'$ will be enlarged, otherwise reduced.
	
	
	The overall linking method is an online greedy algorithm, which is shown in Alg.~\ref{alg:linking}.
	
	\vspace{-0.1in}
	\begin{algorithm}
		\small 
		\renewcommand{\algorithmicrequire}{ \textbf{Input:}}
		\renewcommand{\algorithmicensure}{ \textbf{Output:}}
		\caption{Greedy Linking Algorithm}
		\label{alg:linking}
		\begin{algorithmic}[1]
			\REQUIRE{Predicted Btubes $\{T_i| i \in\{1,2,...,N_T\}\}$}
			\ENSURE{Final tracks $\{K_i | i \in \{1,2,...,N_K\}\}$} 
			
			\STATE Grouping $\{T_i\}$ to $\{H_1, H_2,...,H_L\}$, where $L$ is the total length of the video and \\$H_t = \{T_i^{H_t}| B_m{\rm of}~T^{H_t}_i~{\rm is~at~frame}~t ~\&~ i \in\{1,2,...,N_T\}\}$.
			
			\STATE Utilizing $H_1$ to initialize $\{K_i\}$.
			
			\FOR{$t=2$; $t\leq L$; $t++$ }
			\STATE Calculating $M'$ between $\{K_i\}$ and $H_t$ to form the matching score matrix $S$, where $S_{i,j}=\mathcal{M}'(K_i, T_j^{H_t})$
			
			\STATE Linking the track-tube pairs starting from the largest $S_{i,j}$ in $S$ by Eq.~\ref{eq:link} until all the rest $S_{i,j} <
			\beta $ . Each linking operation will update $\{K_i\}$.
			
			\STATE The remaining Btubes after linking are added to $\{K_i\}$ as new tracks.
			\ENDFOR 
		\end{algorithmic}	
	\end{algorithm}
	\vspace{-0.16in}
	
	\section{Experiments}
	\vspace{-0.1in}
	\paragraph{Datasets and evaluation metrics}
	We  evaluate  our TubeTK model on three MOT Benchmarks~\cite{milan2016mot16,leal2015motchallenge}, namely 2D-MOT2015 (MOT15), MOT16, and MOT17. These benchmarks consist of videos with many occlusions, which makes them really challenging. They are widely used in the field of multi-object tracking and can objectively evaluate models' performances. MOT15 contains 11 train and 11 test videos, while MOT16 and MOT17 contain the same videos, including 7 train and 7 test videos. These three benchmarks provide public detection results (detected by DPM~\cite{felzenszwalb2009object}, Faster R-CNN~\cite{ren2015faster}, and SDP~\cite{yang2016exploit}) for fair comparison among TBD frameworks. However, because our TubeTK conducts MOT in one-step, we do not adopt any external detection results. Without detection results generated by sophisticated detection models trained on large datasets, we need more videos to train the 3D network. Thus, we adopt a synthetic dataset JTA~\cite{fabbri2018learning} which is directly generated from the video game \textit{Grand Theft Auto V} developed by \textit{Rockstar North}. There are 256 video sequences in JTA, enough to pre-train our 3D network. Following the MOT Challenge~\cite{milan2016mot16}, we adopt the CLEAR MOT metrics~\cite{bernardin2008evaluating}, and other measures proposed in~\cite{wu2006tracking}.
	
	\begin{figure*}
		\begin{center}
			\includegraphics[width=\linewidth, height=3.2in]{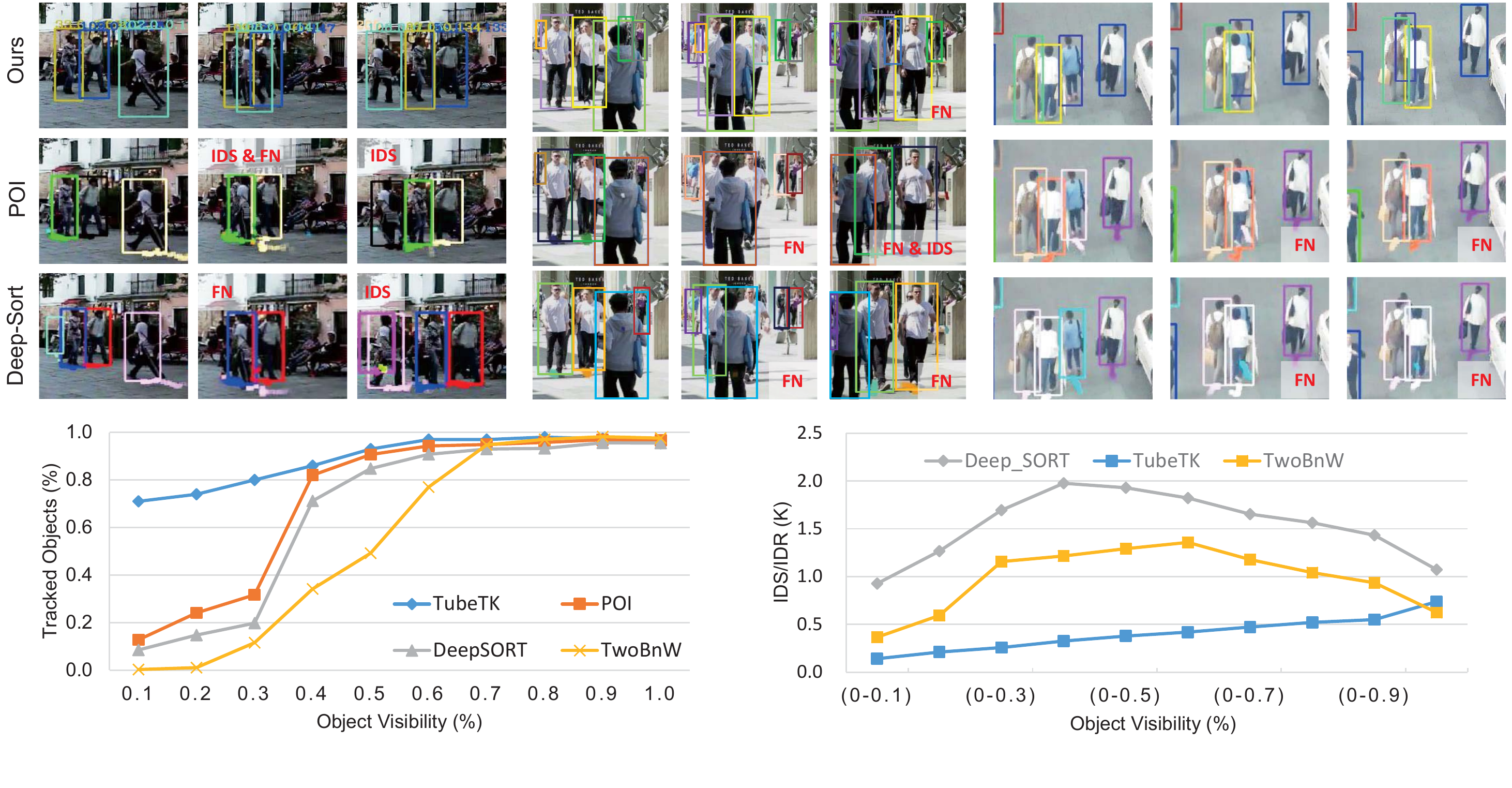}
		\end{center}
		\vspace{-0.15in}
		\caption{Analysis of the performances in occlusion situations. The examples (from test set of MOT16) in the top row show that our TubeTK can effectively reduce the ID switches and false negatives caused by the occlusion. The bottom analysis is conducted on the training set of MOT-16 dataset. We first illustrate the tracked ratio with respect to visibility. The results reveal that our TubeTK performs much better on highly occluded targets than other models. Then, we illustrate the values of IDS/IDR, the conclusion still holds.}
		\label{fig:occlusion}
		\vspace{-0.15in}
	\end{figure*}
	
	\vspace{-0.2in}
	\paragraph{Implementation}
	The hyper-parameters we adopt in the experiments are shown in the following table.
	\begin{table}[H]
		\renewcommand{\arraystretch}{1.2}
		\begin{center}
			\footnotesize
			\vspace{-0.3in}
			\begin{tabular}{ccccccc}
				\hline
				
				\hline
				$\eta$ & $l$ & img size& $\beta$ & $\phi$ & $\gamma_1$ & $\gamma_2$\\
				\hline
				0.8 &  8 & 896$\times$1152 & 0.4 & 0.2 & 0.5 & 0.4\\
				\hline
				
				\hline
			\end{tabular}
		\end{center}
		\label{tab:hy_para}
		\vspace{-0.3in}
	\end{table}
	For each clip $I_t$ we randomly sample a spatial crop from it or its horizontal flip, with the per-pixel mean subtracted. HSL jitter is adopted as color augmentation. The details of the network structure follow FCOS~\cite{tian2019fcos} (see supplementary file for detail). We only replace the 2D CNN layers with the 3D version and modify the last layer in the task head to output tracking results. We initialize the weights as~\cite{tian2019fcos} and train them on JTA from scratch. We utilize SGD with a mini-batch size of 32. The learning rate starts from $10^{-3}$ and is divided by 5 when error plateaus. TubeTK is trained for 150K iterations on JTA and 25K on benchmarks.The weight decay and momentum factors are $10^{-5}$ and 0.9.
	
	\vspace{-0.2in}
	\paragraph{Ablation study}
	The ablation study is conducted on MOT17 training set (without pre-training on JTA). Tab.~\ref{tab:abl_res} demonstrates the great potential of the proposed model. We find that shorter clips ($l=4$) encoding less temporal information lead to bad performance, which reveals that extending the bounding-box to Btube is effective. Moreover, if we fix the length of all the Btubes to 8 (the length of input clips), the performance drops significantly. Fixing length makes the Btubes deviate from the ground-truth trajectory, leading to much more FNs. This demonstrates that setting the length of Btubes dynamically can better capture the moving trails. The other comparisons show the importance of the Tube GIoU loss and Tube NMS. The Original NMS kills many highly occluded Btubes, causing more FN and IDS, and Tube GIoU loss guides the model to regress the Btube's length more accurately than Tube IoU loss (less FN and FP). TubeTK has much more IDS than Tracktor~\cite{bergmann2019tracking} because our FN is much lower and more tracked results potentially lead to more IDS. From IDF1 we can tell that TubeTK tracks better. Note that we refrain from a cross-validation following~\cite{bergmann2019tracking} as our TubeTK is trained on local clips and never accesses to the tracking ground truth data.
	
	\begin{table}[]
		\setlength{\tabcolsep}{1.1mm}
		\caption{Ablation study on the training set of MOT17. D\&T and Tracktor adopt public detections generated by Faster R-CNN~\cite{girshick2015fast}. POI adopts private detection results and is tested on MOT16.}
		\vspace{-0.07in}
		\renewcommand{\arraystretch}{1.2}
		\begin{center}
			\scriptsize
			\begin{tabular}{lccccccc}
				\hline
				
				\hline
				Model & MOTA$\uparrow$ & IDF1$\uparrow$ & MT$\uparrow$ & ML$\downarrow$ & FP$\downarrow$ & FN$\downarrow$ & IDS$\downarrow$\\
				\hline
				\hline
				D\&T~\cite{feichtenhofer2017detect} &50.1&24.9&23.1&27.1&3561&52481&2715\\
				Tracktor++\cite{bergmann2019tracking} &61.9&64.7&35.3&21.4&323&42454&326\\
				POI~\cite{yu2016poi} &65.2&-&37.3&14.7&3497&34241&716\\
				\hline
				TubeTK shorter clips &60.3 & 60.7 & 44.3 & 25.5 & 3446 & 40139 & 968\\
				TubeTK fixed tube len & 74.3 & 68.5 & 62.5 & 8.6 & 7468 & 19452 & 1184\\
				TubeTK IoU Loss & 70.5 & 63.7 & 67.8 & 6.4 & 13247 & 18148 & 1734\\
				TubeTK original NMS & 75.3 & 70.1 & 84.6 & 6.2 & 11256  & 13421 & 2995\\
				TubeTK & 76.9 & 70.0 & 84.7 & 3.1 & 11541 & 11801 & 2687\\
				\hline
				
				\hline
			\end{tabular}
		\end{center}
		\label{tab:abl_res}
		\vspace{-0.3in}
	\end{table}
	
	\vspace{-0.2in}
	\paragraph{Benchmark evaluation}
	
	\begin{table}[]
		\setlength{\tabcolsep}{0.9mm}
		\caption{Results of the online state-of-the-art models on MOT15, 16, 17 datasets. ``Detr" denotes the source of the detection results. Our model does not adopt external detection results (w/o). RAN and CNNMTT utilize the ones provided by POI~\cite{yu2016poi}.}
		\vspace{-0.13in}
		\renewcommand{\arraystretch}{1.2}
		\begin{center}
			\scriptsize
			\begin{tabular}{clcccccccc}
				\hline
				
				\hline
				& Model & Detr& MOTA$\uparrow$ & IDF1$\uparrow$ & MT$\uparrow$ & ML$\downarrow$ & FP$\downarrow$ & FN$\downarrow$ & IDS$\downarrow$\\
				\hline
				\hline
				\multirow{6}{*}{\rotatebox{90}{MOT17}} & Ours & w/o &\textbf{63.0}& \underline{58.6} &\underline{31.2}& \underline{19.9} & 27060 & \underline{177483} & 4137\\
				& SCNet & Priv & \underline{60.0} & 54.4 & \textbf{34.4} & \textbf{16.2} & 72230 & \textbf{145851} & 7611\\
				& LSST17~\cite{feng2019multi} & Pub & 54.7 & \textbf{62.3} & 20.4 & 40.1 & 26091 & 228434 & \textbf{1243}\\
				& Tracktor~\cite{bergmann2019tracking} & Pub & 53.5 & 52.3 & 19.5 & 36.3 & \textbf{12201} & 248047 & \underline{2072}\\
				& JBNOT~\cite{henschel2019multiple} & Pub & 52.6 & 50.8 & 19.7 & 35.8 & 31572 & 232659 & 3050 \\
				& FAMNet~\cite{chu2019famnet} & Pub & 52.0 & 48.7 & 19.1 & 33.4 & \underline{14138} & 253616 & 3072\\
				\hline
				
				\multirow{8}{*}{\rotatebox{90}{MOT16}} & Ours & POI &\textbf{66.9}& 62.2 & \underline{39.0} & \textbf{16.1} & 11544 & \textbf{47502} & 1236\\
				& Ours & w/o & 64.0 & 59.4 & 33.5 & \underline{19.4} & 10962 & 53626 & 1117\\
				& POI~\cite{yu2016poi} & POI & \underline{66.1} & \underline{65.1} & 34.0 & 20.8 & \underline{5061} & 55914 & 805\\
				& CNNMTT~\cite{mahmoudi2019multi} & POI & 65.2 & 62.2 & 32.4 & 21.3 & 6578 & 55896 & 946 \\
				& TAP~\cite{zhou2018online} & Priv & 64.8 & \textbf{73.5} & 38.5 & 21.6 & 12980 & \underline{50635} & \underline{571}\\
				& RAN~\cite{fang2018recurrent} & POI & 63.0 & 63.8 & \textbf{39.9} & 22.1 & 13663 & 53248 & \textbf{482}\\
				& SORT~\cite{bewley2016simple} & Priv & 59.8 & 53.8 & 25.4 & 22.7 & 8698 & 63245 & 1423\\
				& Tracktor~\cite{bergmann2019tracking} & Pub & 54.5 & 52.5 & 19.0 & 36.9 & \textbf{3280} & 79149 & 682\\
				\hline 
				
				\multirow{6}{*}{\rotatebox{90}{MOT15}} & Ours & w/o &\textbf{58.4}& 53.1 & \underline{39.3} & \underline{18.0} & 5756 & \underline{18961} & 854\\
				& RAN~\cite{fang2018recurrent} & POI & \underline{56.5} & \textbf{61.3} & \textbf{45.1} & \textbf{14.6} & 9386 & \textbf{16921} & \underline{428}\\
				& NOMT~\cite{choi2015near} & Priv & 55.5 & \underline{59.1} & 39.0 & 25.8 & \underline{5594} & 21322 & \textbf{427} \\
				& APRCNN~\cite{chen2017online} & Priv & 53.0 & 52.2	& 29.1&	20.2&	\textbf{5159}&	22984& 708\\	
				& CDADDAL~\cite{bae2017confidence} & Priv & 51.3 & 54.1 & 36.3 & 22.2 & 7110 & 22271 & 544\\
				& Tracktor~\cite{bergmann2019tracking} & Pub & 44.1 & 46.7 & 18.0 & 26.2 & 6477 & 26577 & 1318\\
				\hline
				
				\hline
			\end{tabular}
		\end{center}
		\label{tab:compare_res}
		\vspace{-0.35in}
	\end{table}
	
	Tab.~\ref{tab:compare_res} presents the results of our TubeTK and other state-of-the-art (SOTA) models which adopt public or private external detection results (detailed results are shown in  supplementary files). We only compare with the officially published and peer-reviewed online models in the MOT Challenge benchmark\footnote{MOT challenge leaderboard: \url{https://motchallenge.net}}. As we show, although TubeTK does not adopt any external detection results, it achieves new SOTA results on MOT17 (3.0 MOTA improvements) and MOT15 (1.9 MOTA improvements). On MOT16, it achieves much better performance than other SOTAs that rely on publicly available detections (64.0 vs. 54.5). Moreover, TubeTK performs competitively with the SOTA models adopting POI~\cite{yu2016poi} detection bounding-boxes and appearance features (POI-D-F)\footnote{\url{https://drive.google.com/open?id=0B5ACiy41McAHMjczS2p0dFg3emM}} on MOT16.
	It should be noted that the authors of POI-D-F utilize 2 extra tracking datasets, many self-collected surveillance data (10$\times$ frames than MOT16) to train the Faster-RCNN detector, and 4 extra Re-ID datasets to extract the appearance features. Thus, we cannot get the same generalization ability as the POI-D-F with synthetic JTA data. To demonstrate the potential of TubeTK, we also provide the results adopting the POI detection (without the appearance features, details in supplementary files) and in this setting our TubeTK achieves the new state-of-the-art on MOT16 (66.9 vs. 66.1). On these three benchmarks, due to the great resistibility to occlusions, our model has fewer FN, under the condition that the number of FP is relatively acceptable. Although TubeTK can handle occlusions better, its IDS is relatively higher because we do not adopt feature matching mechanisms to maintain global consistency. The situation of IDS in occlusion parts is further discussed in Sec.~\ref{sec:occlu}.
	
	
	\vspace{-0.05in}
	\section{Discussion}
	\vspace{-0.05in}
	\paragraph{Overcoming the occlusion}~\label{sec:occlu}
	With Btubes, our model can learn and encode the moving trend of targets, leading to more robust performances when facing severe occlusions. We show the qualitative and quantitative analysis in Fig.~\ref{fig:occlusion}. Form the top part of Fig.~\ref{fig:occlusion}, we show that TubeTK can keep tracking with much less FN or IDS when the target is totally shielded by other targets. In the bottom part, we provide the tracked ratio and number of IDS (regularized by ID recall) with respect to targets' visibility on the training set of MOT16. When the visibility is low, TubeTK performs much better than other TBD models.
	
	\vspace{-0.15in}
	\paragraph{Robustness of Btubes for linking}
	The final linking process has no learnable parameters, thus the linking performances depend heavily on the accuracy of regressed Btubes. To verify the robustness, we perform the linking algorithm on GT Btubes with noise jitter. The jitter is conducted on Btubes' center position and spatial-temporal scale. 0.25 jitter on center position or scale means the position or scale shift up to 25\% of the Btube's size. The results on MOT17-02, a video with many crossovers, are shown in Tab.~\ref{tab:robust}. We can find that even with large jitter up to 25\%, the linking results are still great enough (MOTA $>$ 86, IDF1 $>$ 79), which reveals that the linking algorithm is robust and does not need rigorously accurate Btubes to finish the tracking.
	\begin{table}[]
		\setlength{\tabcolsep}{0.8mm}
		\caption{Experiments on linking robustness. We only test on the GT tracks of a single video MOT17-02. ``cn" and ``sn" denote the center position and bounding-box scale noises. In each grid, the values are ``MOTA" ``IDF1", ``MT", and ``ML" in order.}
		\vspace{-0.1in}
		\renewcommand{\arraystretch}{1.2}
		\newcommand{\tabincell}[2]{\begin{tabular}{@{}#1@{}}#2\end{tabular}}
		\begin{center}
			\scriptsize
			\begin{tabular}{c||cc|cc|cc|cc|cc|cc}
				\hline
				
				\hline
				\diagbox{cn}{sn} & \multicolumn{2}{c|}{0.00} & \multicolumn{2}{c|}{0.05} & \multicolumn{2}{c|}{0.10} &  \multicolumn{2}{c|}{0.15} & \multicolumn{2}{c|}{0.20}& \multicolumn{2}{c}{0.25}\\
				\hline
				\hline
				0.00 & \tabincell{c}{97.2\\59} & \tabincell{c}{91.5\\0} & \tabincell{c}{95.2\\58} & \tabincell{c}{91.3\\0} & \tabincell{c}{95.0\\58} & \tabincell{c}{91.2\\0}& \tabincell{c}{94.3\\58} & \tabincell{c}{91.4\\0} &
				\tabincell{c}{93.6\\54} & \tabincell{c}{91.4\\0} &
				\tabincell{c}{92.4\\53} & \tabincell{c}{86.2\\1}\\
				\hline
				0.05 & \tabincell{c}{96.1\\58} & \tabincell{c}{91.5\\0} & \tabincell{c}{95.2\\57} & \tabincell{c}{90.8\\0} & \tabincell{c}{95.1\\58} & \tabincell{c}{91.5\\0}& \tabincell{c}{95.9\\58} & \tabincell{c}{91.3\\0} &
				\tabincell{c}{94.9\\58} & \tabincell{c}{91.9\\0} &
				\tabincell{c}{96.5\\59} & \tabincell{c}{89.6\\0}\\
				\hline
				0.15 & \tabincell{c}{94.2\\56} & \tabincell{c}{89.1\\0} & \tabincell{c}{94.4\\54} & \tabincell{c}{91.8\\2} & \tabincell{c}{95.1\\56} & \tabincell{c}{89.4\\1}& \tabincell{c}{96.3\\56} & \tabincell{c}{91.3\\1} &
				\tabincell{c}{94.3\\56} & \tabincell{c}{87.4\\0} &
				\tabincell{c}{94.0\\55} & \tabincell{c}{91.3\\0}\\
				\hline
				0.25 & \tabincell{c}{91.6\\54} & \tabincell{c}{84.7\\2} & \tabincell{c}{91.1\\55} & \tabincell{c}{81.9\\2} & \tabincell{c}{92.8\\54} & \tabincell{c}{83.1\\3}& \tabincell{c}{87.8\\54} & \tabincell{c}{82.8\\2} &
				\tabincell{c}{88.5\\53} & \tabincell{c}{83.4\\2} &
				\tabincell{c}{86.4\\54} & \tabincell{c}{79.9\\2}\\
				\hline
				
				\hline
			\end{tabular}
		\end{center}
		\label{tab:robust}
		\vspace{-0.3in}
	\end{table}
	
	\vspace{-0.05in}
	\section{Conclusion}
	\vspace{-0.05in}
	In this paper, we proposed an end-to-end one-step training model TubeTK for MOT task. It utilizes Btubes to encode target's temporal-spatial position and local moving trail. This makes the model independent of external detection results and has enormous potential to overcome occlusions. We conducted extensive experiments to evaluate the proposed model. On the mainstream benchmarks, our model achieves the new state-of-the-art performances compared with other online models, even if they adopt private detection results. Comprehensive analyses were presented to further validate the robustness of TubeTK.
	
	\vspace{-0.05in}
	\section{Acknowledgements}
	\vspace{-0.05in}
	This work is supported in part by the National Key R\&D Program of China, No. 2017YFA0700800, National Natural Science Foundation of China under Grants 61772332 and Shanghai Qi Zhi Institute.
	
	{\small
		\bibliographystyle{ieee_fullname}
		\bibliography{egbib}
	}
	
\end{document}